\documentclass{article}

\newif\ifneuripsstyle
\IfFileExists{neurips_2026.sty}{
    \neuripsstyletrue
    \usepackage[preprint]{neurips_2026}
}{
    \neuripsstylefalse
    \usepackage[margin=1in]{geometry}
    \usepackage{times}
    \usepackage[numbers]{natbib}
}

\usepackage{amsmath,amssymb,amsthm}
\usepackage{booktabs}
\usepackage{graphicx}
\usepackage{hyperref}
\usepackage{xcolor}
\usepackage{enumitem}
\usepackage{microtype}
\usepackage{bbm}
\usepackage{algorithm}
\usepackage{algpseudocode}

\title{TREK: Distill to Explore, Reinforce to Refine}

\author{
    Yuanda Xu$^{1}$\thanks{Equal contribution.}\thanks{Correspondence to \texttt{yuanda@math.princeton.edu}} \quad
    Zhengze Zhou$^{1}$\footnotemark[1] \quad
    Kayhan Behdin$^{1}$ \quad
    Jelena Markovic-Voronov$^{1}$ \\
    {\bf Hejian Sang$^{1}$\thanks{Work done while at LinkedIn Corporation.} \quad
    Xiaomin Li$^{2}$ \quad Wenhui Zhu$^{1}$ \quad Xinchen Du$^{1,3}$\thanks{Work done while an intern at LinkedIn Corporation.} \quad Aida Rahmattalabi$^{1}$ } \\
    {\bf Ran He$^{1}$ \quad
    Sen Na$^{3}$ \quad
    Zhipeng Wang$^{1}$\footnotemark[3] \quad
    Alborz Geramifard$^{1}$} \\
    \vspace{0.2cm}
    $^{1}$LinkedIn Corporation \quad
    $^{2}$Harvard University \quad
    $^{3}$Georgia Institute of Technology
}

\date{}

\newcommand{\student}{\pi_{\theta}}
\newcommand{\studentold}{\pi_{\theta_{\mathrm{old}}}}
\newcommand{\teacher}{\pi_{T}}
\newcommand{\verifier}{V}
\newcommand{\ps}{p_S}
\newcommand{\nll}{d_S}
\newcommand{\segrpo}{TREK}
\newcommand{\Yreach}{\mathcal{Y}_{\mathrm{reach}}}
\newcommand{\Dscaf}{\mathcal{D}_{\mathrm{scaf}}}

\begin{document}
\maketitle

\begin{abstract}
Group Relative Policy Optimization (GRPO) is effective when the current policy already samples useful reasoning trajectories, but it stalls on hard prompts whose correct solution modes lie outside the student's on-policy support. We propose \segrpo{} (\textbf{T}eacher-\textbf{R}outed \textbf{E}xploration via Forward \textbf{K}L), a simple staged procedure that uses distillation not for imitation but for \emph{exploration support expansion}. A key advantage of \segrpo{} is its generality: because it only consumes verified output trajectories, it can use an external black-box teacher, a white-box teacher, or the same model given additional inference-time context, and it can efficiently identify which hard-prompt samples are most worth consolidating even when teacher internals are unavailable. \segrpo{} first identifies prompts where the unaided student has very low pass rate, queries a proposal source to produce verified candidate solutions, keeps the top-$r$ proposals ranked by current student likelihood, applies a short forward-KL phase to pull those verified modes into the student's support, and then returns to standard on-policy GRPO refinement. On mathematical reasoning, \segrpo{} with DeepSeek-V4 proposals improves Qwen3 models across all tested scales on AIME 2024 and AIME 2025; for Qwen3-8B, it improves AIME 2025 from $36.9$ to $40.3$ and AIME 2024 from $47.9$ to $51.1$ (avg@16), while the self-context variant reaches $38.5$ and $49.6$ without an external teacher. On agentic tasks, \segrpo{} raises ALFWorld success rate from $75.8$ to $82.8$ and ScienceWorld success rate from $12.5$ to $26.7$; notably, on the hardest task types, \segrpo{} achieves high success rates early in training while unaided GRPO requires substantially more optimization steps to reach comparable levels.
\end{abstract}

\section{Introduction}

Distillation, curriculum learning, and reinforcement learning from verifiable rewards have become central tools for improving language-model reasoning \citep{hinton2015distilling,kim2016sequence,bengio2009curriculum,kumar2010selfpaced,shao2024deepseekmath,deepseek2025r1}. In Group Relative Policy Optimization (GRPO), a policy samples a group of candidate responses for each prompt and receives relative advantages from outcome rewards or programmatic verifiers. The appeal of GRPO is that it directly optimizes the behaviors that the deployed model can sample, avoiding a learned value model and keeping training closely tied to test-time behavior.

The same on-policy property also creates a limitation. GRPO is most effective when the current student already assigns nontrivial probability mass to useful solution modes. On harder prompts, however, the student may repeatedly sample plausible but structurally similar wrong trajectories. In this regime, the verifier is not the scarce resource: it can score a correct solution if one appears. The scarce resource is exploration coverage of the solution space. Larger rollout groups or sharper relative advantages may still search within a narrow region of the student's current support.

This paper studies a more concrete question: \emph{how can we improve GRPO when the bottleneck is not reward sparsity on sampled trajectories, but inadequate exploration beyond the student's current support?} Recent teacher-informed and on-policy distillation methods largely improve supervision over trajectories the student already reaches, for example by measuring teachability, adapting supervision to student competence, weighting tokens, improving credit assignment, iteratively distilling self-policy behavior, or smoothing optimization on sparse rewards \citep{agarwal2024onpolicy,gu2024minillm,wang2026teachability,xu2026paced,xu2026tip,shen2026credit,sang2026beyond, sang2026crisp}. These directions are complementary, but they do not directly address prompts on which the student rarely samples any reward-bearing trajectory. In that regime, the first problem is not how to assign better credit to sampled behavior; it is how to make useful behavior sampleable at all.

We therefore study \emph{distillation as exploration}. The central idea is that a teacher or context-augmented model should be treated as a proposal mechanism, not as a universal behavior target. A stronger model, or the same model given additional inference-time context such as execution feedback, reflection, a longer reasoning budget, or failure lessons, can discover verified solutions that the unaided student currently misses \citep{wang2022selfconsistency,zelikman2022star,yao2023react,shinn2023reflexion,yao2023tree}. This design deliberately goes beyond white-box teacher-dependent recipes: the proposal source may be an external black-box or white-box teacher, or the same model run with additional context, because the method needs verified trajectories rather than teacher logits, probabilities, hidden states, or other internal signals. This output-only interface makes the method broadly applicable in practical black-box settings, where one can query a capable system but cannot inspect its probabilities or activations. It also turns the proposal source into an efficient sample-selection engine: among many possible verified successes, \segrpo{} keeps the samples that are closest to the current student and therefore most useful for expanding the student's own sampling support. The question is how to convert those proposals into improved \emph{unaided} exploration. Our answer is forward-KL proposal learning: if the teacher discovers verified solution modes that are novel but still reachable under the current student, then minimizing $\mathrm{KL}(q_{\mathrm{prop}}\|\student)$ increases the student's probability mass on those modes, making them available to later on-policy RL.

\begin{figure}[t]
    \centering
    \includegraphics[width=0.98\linewidth]{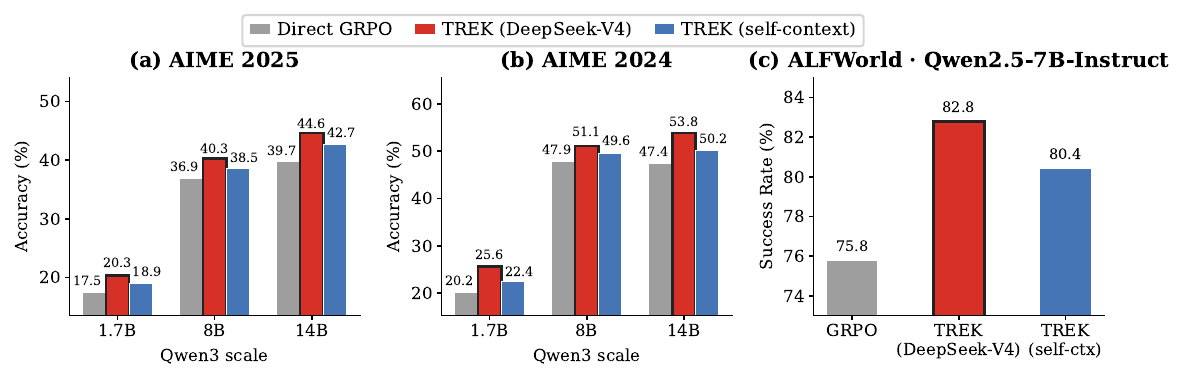}
    \caption{Summary of main results. (a)--(b): Math reasoning accuracy (avg@16, \%) across Qwen3 scales on AIME 2025 and AIME 2024. \segrpo{} consistently improves over direct GRPO at all model sizes, with the DeepSeek-V4 proposal source outperforming the self-context variant. (c): ALFWorld success rate with Qwen2.5-7B-Instruct, showing analogous gains in the agentic planning domain. ScienceWorld results are reported in Table~\ref{tab:alfworld-final-results}.}
    \label{fig:main-results}
\end{figure}

We instantiate this view in \segrpo{}, a simple routed procedure that separates \emph{prompt-level proposal availability} from \emph{trajectory-level reachability}. The method identifies hard prompts using unaided student pass rate, invokes a proposal source only on those prompts, keeps top-$r$ verified proposals by current student likelihood, applies a short forward-KL warm-start on those trajectories, and then returns control to the ordinary GRPO rollout stream. The proposal source suggests where the student should search next; the verifier and GRPO determine what should ultimately be reinforced.

Our contributions are:
\begin{enumerate}[leftmargin=*]
    \item We identify missing exploration support as a distinct failure mode of GRPO on hard reasoning prompts, and show how an output-only proposal interface makes targeted support expansion compatible with black-box teachers, white-box teachers, and same-model context augmentation.
    \item We formulate a broader view of distillation as exploration support expansion, rather than only imitation or token-level credit shaping.
    \item We instantiate this view in \segrpo{}, a simple routed procedure that explicitly separates prompt-level proposal availability from trajectory-level reachability, then uses short forward-KL consolidation before subsequent on-policy GRPO refinement.
    \item On agentic tasks, we show that \segrpo{} not only improves final success rate but significantly accelerates early training on the hardest task types: by injecting verified solution modes before RL begins, the student achieves high success rates early in training while unaided GRPO requires substantially more steps to reach comparable levels.
\end{enumerate}

\section{Method}

\begin{figure}[!ht]
    \centering
    \includegraphics[width=0.98\linewidth]{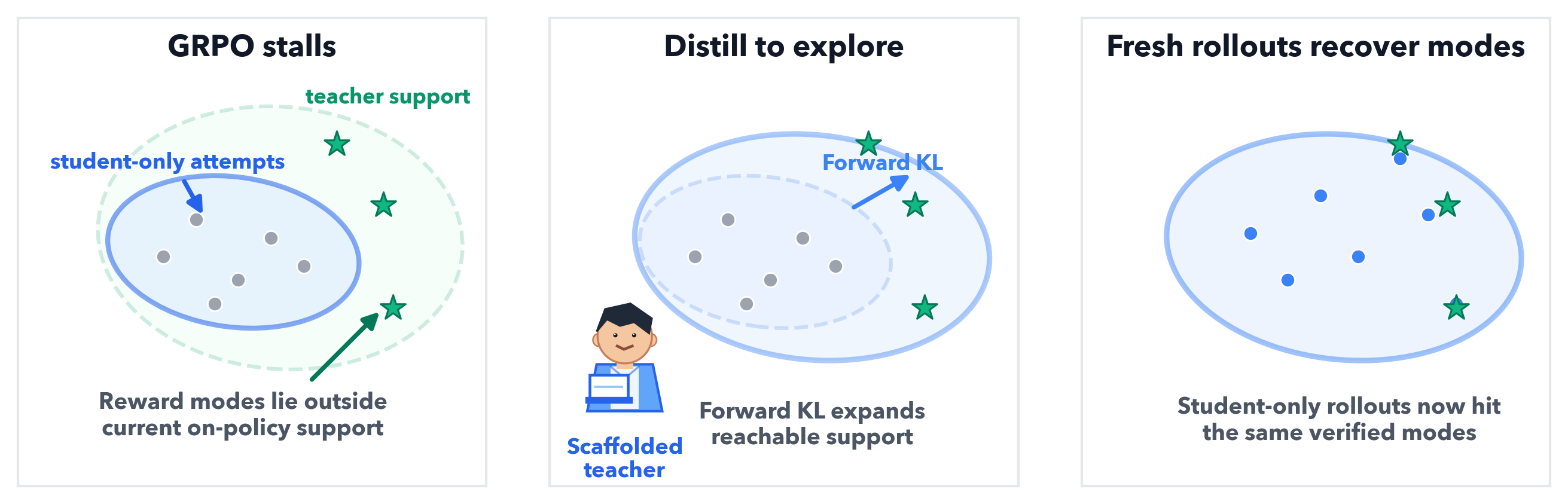}
    \caption{Overview of \segrpo{} as distillation for exploration support expansion. Standard GRPO stalls when student-only attempts remain inside a narrow on-policy support while verified reward modes lie outside it. A proposal source has broader support, either from a stronger model or from the same model with additional context, but only top-$r$ student-proximal verified modes are consolidated; a short forward-KL update expands the student's reachable support toward those modes. The key test is then sample-level recovery: fresh student-only rollouts should reach the same verified modes, after which ordinary GRPO can optimize them using verifier rewards.}
    \label{fig:trek-overview}
\end{figure}

\subsection{Problem Setup and Failure Mode}

Let $x$ denote a prompt, $\student(\cdot\mid x)$ the student policy, $\teacher(\cdot\mid x)$ a proposal policy, and $\verifier(y)\in\{0,1\}$ a verifier that checks whether a response or trajectory $y$ succeeds. Standard GRPO optimizes the student from responses sampled from its current policy. Writing $\{y_i\}_{i=1}^{G}\sim \student(\cdot\mid x)$ for a rollout group and $r_i=\verifier(y_i)$ for rewards, the group-normalized advantage is
\begin{equation}
    A_i = \frac{r_i - \mathrm{mean}(\{r_j\}_{j=1}^{G})}{\mathrm{std}(\{r_j\}_{j=1}^{G}) + \epsilon},
\end{equation}
and the corresponding clipped objective can be written as \citep{schulman2017proximal}
\begin{equation}
    \mathcal{L}_{\mathrm{GRPO}}(x)
    = -\frac{1}{G}\sum_{i=1}^{G}\sum_{t=1}^{|y_i|}
    \min\left(\rho_{i,t} A_i, \mathrm{clip}(\rho_{i,t},1-\epsilon,1+\epsilon) A_i\right)
    + \beta\,\mathrm{KL}(\student \| \pi_{\mathrm{ref}}),
\end{equation}
where $\rho_{i,t}=\student(y_{i,t}\mid x,y_{i,<t})/\studentold(y_{i,t}\mid x,y_{i,<t})$. The key limitation for this paper is that GRPO can only reinforce modes that the student already samples.

We target prompts where that assumption fails. The student's unaided pass rate under the training-time decoding configuration is
\begin{equation}
    \ps(x)=\frac{1}{K}\sum_{i=1}^{K}\mathbbm{1}[\verifier(y_S^i)=1], \quad y_S^i\sim \student(\cdot\mid x),
\end{equation}
where $y_S^i$ are ordinary student rollouts. Prompts with low $\ps$ are hard candidates: the unaided student fails to discover useful trajectories under the current sampling configuration. A hard candidate becomes useful for proposal learning only if the proposal source later produces at least one verified trajectory. This is the setting in which we treat distillation as exploration support expansion rather than denser supervision on already-sampled rollouts.

For a verified teacher trajectory $y_T$, let
\begin{equation}
    \ell_t(y_T\mid x)=-\log \student(y_{T,t}\mid x,y_{T,<t})
\end{equation}
be its token-level student NLL. We measure reachability with a two-sided trimmed length-normalized NLL,
\begin{equation}
    \nll(y_T\mid x)=\frac{1}{|\mathcal{I}_{\alpha,\beta}(y_T)|}\sum_{t\in\mathcal{I}_{\alpha,\beta}(y_T)}\ell_t(y_T\mid x),
\end{equation}
where $\mathcal{I}_{\alpha,\beta}(y_T)$ removes the lowest $\alpha$ fraction and highest $\beta$ fraction of token losses before averaging. The low-end trim prevents high-confidence boilerplate or formatting tokens from making a trajectory look artificially close, while the high-end trim prevents isolated rare-token outliers from making it look artificially far. The resulting quantity is a \emph{reachability score}, not yet a training loss: smaller $\nll(y_T\mid x)$ indicates a verified trajectory closer to the student's current support. Because absolute NLL scales differ across prompts, we also use the prompt-relative form
\begin{equation}
    \Delta d(y_T\mid x) = \nll(y_T\mid x) - \mathbbm{E}_{y\sim \student(\cdot\mid x)}[\nll(y\mid x)]
\end{equation}
as an analysis variable, where the expectation is estimated from the same student rollouts already drawn for $\ps(x)$. We deliberately do not commit to fixed absolute thresholds on $\nll$ or $\Delta d$ in the main method: selection is governed by a ranking rule defined below, and $\nll$, $\Delta d$ are reused only as diagnostics for bucketed analysis.

\subsection{Prompt Routing and Reachability}

\segrpo{} first decides where proposal generation is worth paying for. For each prompt $x$, we estimate the unaided student pass rate $\ps(x)$ and query the proposal source only on hard prompts,
\begin{equation}
\text{Routed hard prompt:}\quad \ps(x) \le \tau_{\mathrm{low}}.
\end{equation}
Prompts above this cutoff remain in the ordinary GRPO stream. A hard prompt receives proposal learning only if the proposal source later produces verified trajectories that are close enough to the current student to consolidate. Concrete values of $K$, $M$, and $\tau_{\mathrm{low}}$ are given in Appendix~\ref{app:setup-details}.

The proposal policy $\teacher$ is only used to generate candidate solutions for hard prompts. It may be a larger model, but the most deployment-aligned setting is the same student family run with additional inference-time context. Examples include verifier-guided retry, self-consistency, search, environment interaction, reflection, failure lessons, or a longer reasoning budget. This extra context or computation may help discover solutions, but retained targets must still be expressible as ordinary student outputs at deployment time.

For each routed hard prompt, we discard proposal trajectories that fail the verifier. Among the verified trajectories, we rank by current student likelihood and keep the top-$r$ \emph{student-proximal verified proposals},
\begin{equation}
    \Yreach(x)=\operatorname*{Top}_{r}\,\bigl\{y_T:\verifier(y_T)=1\bigr\}\ \text{by increasing}\ \nll(y_T\mid x).
\end{equation}
This rule makes proposal learning conditional on teacher success while avoiding broad imitation of trajectories far from the student's current support. The hyperparameter $r$ trades off multi-mode coverage against transfer stability. We use trimmed NLL for selection and reuse $\nll$ and $\Delta d$ only as analysis variables, not as extra thresholds.

\subsection{Proposal Learning and On-Policy Refinement}

On routed hard prompts, $\Yreach(x)$ defines the target for a short support-expansion update. If teacher probabilities are available, the retained proposal distribution can be written as
\begin{equation}
    q_{\mathrm{prop}}(y\mid x) \propto \teacher(y\mid x)\,\mathbbm{1}[\verifier(y)=1] \,\mathbbm{1}[y\in \Yreach(x)].
\end{equation}
Explicit teacher probabilities are not required; retained samples are sufficient. At the distributional level, proposal learning minimizes
\begin{equation}
    \mathcal{L}_{\mathrm{FKL}}(x)=\mathrm{KL}(q_{\mathrm{prop}}(\cdot\mid x)\|\student(\cdot\mid x))
    = \mathbbm{E}_{y\sim q_{\mathrm{prop}}}\left[-\log \student(y\mid x)\right] + \mathrm{const.}
\end{equation}
on selected prompts. The forward direction penalizes missing proposal support and encourages the student to cover the retained solution modes. In black-box or context-only settings, this reduces to teacher-forced negative log-likelihood on retained proposal samples,
\begin{equation}
    \mathcal{L}_{\mathrm{prop}}(x,y_T) = -\sum_{t=1}^{|y_T|}\log \student(y_{T,t}\mid x,y_{T,<t}).
\end{equation}
We apply this SFT-like update only on the selected proposal dataset
\begin{equation}
    \Dscaf = \{(x,y_T): \ps(x)\leq \tau_{\mathrm{low}},\ y_T\in \Yreach(x)\},
\end{equation}
which contains only verified, top-$r$ student-proximal trajectories from prompts that the unaided student currently solves only rarely.

The overall schedule is staged rather than a single mixed minibatch:
\begin{equation}
    \text{hard prompt mining} \rightarrow \text{proposal selection} \rightarrow \text{short forward-KL warm-start} \rightarrow \text{ordinary GRPO refinement}.
\end{equation}
Hard prompts without retained verified proposals simply remain in the GRPO stream until the next routing refresh. Appendix~\ref{app:pipelined-schedule} describes an optional pipelined schedule that overlaps ordinary GRPO with delayed teacher proposal generation.

\subsection{Algorithm}

\begin{algorithm}[t]
\caption{TREK}
\label{alg:trek}
\begin{algorithmic}[1]
\State Sample a batch of prompts $\mathcal{B}$ and initialize $\Dscaf\gets\emptyset$.
\For{each $x\in\mathcal{B}$}
    \State Sample $K$ unaided student rollouts and estimate $\ps(x)$.
    \If{$\ps(x)\le \tau_{\mathrm{low}}$}
        \State Generate up to $M$ proposal rollouts.
        \State Keep verifier-passing teacher trajectories.
        \State Rank retained trajectories by $\nll(y_T\mid x)$ and keep the top-$r$ as $\Yreach(x)$.
        \If{$\Yreach(x)\neq\emptyset$}
            \State $\Dscaf\gets\Dscaf\cup\{(x,y_T):y_T\in\Yreach(x)\}$.
        \Else
            \State Defer $x$ until the next routing refresh.
        \EndIf
    \Else
        \State Route $x$ to ordinary GRPO and revisit periodically.
    \EndIf
\EndFor
\If{$\Dscaf\neq\emptyset$}
    \State Apply a short forward-KL proposal-learning phase on the accumulated $\Dscaf$.
    \State Return proposal-updated prompts to the normal sampling pool for later GRPO rollouts.
\EndIf
\State Periodically refresh $\ps$, verified-proposal availability, and reachability statistics.
\end{algorithmic}
\end{algorithm}

The paper focuses on this minimal instantiation. More adaptive schedulers and per-prompt stopping rules are possible extensions, but they are not required to test the main mechanism. The refresh interval controls a tradeoff between routing accuracy and overhead: early in training, frequent refreshes are useful because prompts can move into or out of the low-pass-rate hard set, while later in training cached routing statistics can be reused for longer windows.

\section{Experiments}

We evaluate whether \segrpo{} improves over ordinary GRPO on hard prompts, whether those gains survive budget-matched comparisons, and whether ablations support the proposed routing, proposal-selection, and consolidation mechanisms. Figure~\ref{fig:main-results} summarizes the main results across math and ALFWorld; the rest of this section gives the detailed numbers and ablations.

\subsection{Experimental Setup}

For mathematical reasoning, we evaluate Qwen3 deployment models at 1.7B, 8B, and 14B scales \citep{yang2025qwen3}. The reported \segrpo{} rows use verified rollouts generated by DeepSeek-V4 as the proposal teacher \citep{deepseek2026v4}, then consolidate the selected trajectories back into the Qwen3 deployment model. We also include a separate \segrpo{} self-context variant, where the proposal source is the same Qwen3 model run with additional inference-time context; concretely, the prompt is augmented with a short failure-lesson memory self-summarized from the model's previously verifier-rejected attempts (Appendix~\ref{app:self-context}). For ALFWorld, the deployable student model is Qwen2.5-7B-Instruct \citep{qwen2024qwen25}.

Math benchmarks are AIME 2024 and AIME 2025, with exact-match or symbolic answer checking. All math numbers are reported as avg@16: the per-problem unaided pass rate is estimated from $16$ independent samples drawn from the deployment student under the same decoding configuration used at training time, then averaged over the benchmark. For ALFWorld \citep{shridhar2021alfworld}, each prompt is an initial observation together with the task instruction, a trajectory is a full observation--action transcript, and the verifier is the environment success signal. During proposal generation, the proposal source may use a larger model or additional inference-time context and computation such as self-consistency, verifier-guided retry, repair loops, admissible-action filtering, environment feedback, or backtracking; the stored consolidation target is always the final verified response or trajectory. Final evaluation always uses the unaided deployment student without tools, retry, or extra planning context. Because agentic training has high run-to-run variance, we aggregate ALFWorld and ScienceWorld results over ten independent end-to-end training-and-evaluation runs and report mean $\pm$ sample standard deviation rather than relying on a single seed.

We keep the main recipe intentionally simple: in every training round, the current student is rolled out a small number of times per prompt to estimate $\ps(x)$; on prompts with low estimated unaided pass rate, the proposal source is queried a handful of times, verifier-passing trajectories are kept, and the top-$r$ student-proximal ones are accumulated into $\Dscaf$ for a short proposal-learning phase before the prompts return to the ordinary GRPO sampling stream. The concrete sample sizes for $K$ and $M$, the hardness cutoff $\tau_{\mathrm{low}}$, the selection size $r$, and the proposal-learning window length are reported in Appendix~\ref{app:setup-details}.

\subsection{Math Results}

Table~\ref{tab:qwen3-math-results} reports direct GRPO performance and \segrpo{} performance across Qwen3 scales. The direct-GRPO rows are deployment-student baselines. The reported \segrpo{} rows use DeepSeek-V4 verified rollouts as external black-box proposals; the self-context rows appended at the bottom test the matched same-model context variant. We focus the main tables on the two AIME benchmarks, which probe the hard-prompt exploration regime; the corresponding MATH-500 numbers \citep{hendrycks2021measuring}, where direct GRPO is already near-saturated and gains are correspondingly small, are reported in Appendix~\ref{app:math500}.

\begin{table}[!ht]
\centering
\begin{tabular}{llcc}
\toprule
Stage & Model & AIME 2024 & AIME 2025 \\
\midrule
Direct GRPO & Qwen3-1.7B & $20.2 \pm 1.3$ & $17.5 \pm 1.0$ \\
Direct GRPO & Qwen3-8B & $47.9 \pm 1.4$ & $36.9 \pm 1.2$ \\
Direct GRPO & Qwen3-14B & $47.4 \pm 1.1$ & $39.7 \pm 0.9$ \\
\midrule
\segrpo{} (DeepSeek-V4) & Qwen3-1.7B & $25.6 \pm 1.5$ & $20.3 \pm 1.2$ \\
\segrpo{} (DeepSeek-V4) & Qwen3-8B & $51.1 \pm 1.6$ & $40.3 \pm 1.4$ \\
\segrpo{} (DeepSeek-V4) & Qwen3-14B & $53.8 \pm 1.4$ & $44.6 \pm 1.2$ \\
\midrule
\segrpo{} (self-context) & Qwen3-1.7B & $22.4 \pm 1.2$ & $18.9 \pm 1.2$ \\
\segrpo{} (self-context) & Qwen3-8B & $49.6 \pm 1.5$ & $38.5 \pm 1.1$ \\
\segrpo{} (self-context) & Qwen3-14B & $50.2 \pm 1.2$ & $42.7 \pm 1.0$ \\
\midrule
OPD (self-context) & Qwen3-1.7B & $21.3 \pm 1.3$ & $16.0 \pm 1.1$ \\
OPD (self-context) & Qwen3-8B & $48.4 \pm 1.3$ & $36.7 \pm 0.9$ \\
OPD (self-context) & Qwen3-14B & $48.6 \pm 1.2$ & $40.2 \pm 1.0$ \\
\bottomrule
\end{tabular}
\vspace{4pt}
\caption{Math results across Qwen3 scales on AIME 2024 and AIME 2025 (avg@16, \%). The reported \segrpo{} results use verified DeepSeek-V4 rollouts as proposal trajectories, then consolidate the selected trajectories into the Qwen3 deployment model through a short forward-KL proposal-learning phase before GRPO refinement resumes. The self-context rows are the matched same-model context variant. The OPD rows use the same self-context proposal source but replace forward-KL consolidation with on-policy distillation-style supervision, showing that the forward-KL objective provides a stronger support-expansion signal for that matched proposal source.}
\label{tab:qwen3-math-results}
\end{table}

The main pattern in Table~\ref{tab:qwen3-math-results} is that support-expansion proposals help consistently on the harder AIME benchmarks. With DeepSeek-V4 proposals, \segrpo{} improves every Qwen3 scale over direct GRPO: Qwen3-1.7B gains $+5.4$ points on AIME 2024 and $+2.8$ on AIME 2025, Qwen3-8B gains $+3.2$ and $+3.4$, and Qwen3-14B gains $+6.4$ and $+4.9$. This matches the intuition that proposal learning is most useful when unaided sampling misses reward-bearing modes, exactly the regime that the harder AIME benchmarks probe. The self-context rows show that the same mechanism does not require an external teacher: adding inference-time context to the deployment model itself still improves over direct GRPO at every scale, reaching $49.6$/$38.5$ for Qwen3-8B and $50.2$/$42.7$ for Qwen3-14B on AIME 2024/2025. Replacing forward-KL consolidation with OPD-style supervision weakens the matched self-context variant across all scales: OPD trails \segrpo{} (self-context) by $1.1$--$1.6$ points on AIME 2024 and by $1.8$--$2.9$ points on AIME 2025. The gap is interpretable: forward KL directly penalizes the student for assigning low probability to verified proposal modes, so the gradient explicitly pushes mass toward those modes; OPD-style supervision instead shapes credit on student-sampled trajectories, but on hard prompts where the student rarely reaches reward-bearing modes in the first place, there is little useful on-policy signal for OPD to redistribute. Even when OPD is applied off-policy on the same teacher trajectories, it lacks the explicit missing-support penalty of forward KL and tends to match surface form rather than expand mode coverage. This supports the design choice of forward-KL as the consolidation objective for support expansion on hard prompts.

All methods are matched on student optimization steps. The final evaluation always uses the unaided student, without tools, verifier-guided retry, or teacher assistance. The compact tables report the primary equal-student-update comparisons across proposal sources in each setting.

\subsection{Agentic Task Results}

Table~\ref{tab:alfworld-final-results} compares direct GRPO with the external DeepSeek-V4 and self-context proposal sources on two agentic benchmarks: ALFWorld \citep{shridhar2021alfworld}, a household planning environment, and ScienceWorld \citep{wang2022scienceworld}, a text-based science experiment simulator requiring multi-step procedural reasoning. Table~\ref{tab:alfworld-bytask} reports the accompanying ALFWorld per-task-type breakdown that most directly tests our central claim. The per-task breakdown shows \emph{where} the gains land: the lowest-baseline task types are exactly the ones \segrpo{} improves most.

\begin{table}[t]
\centering
\begin{tabular}{lcc}
\toprule
Method & ALFWorld & ScienceWorld \\
\midrule
GRPO & $75.8 \pm 2.1$ & $12.5 \pm 1.7$ \\
\segrpo{} (DeepSeek-V4) & $82.8 \pm 2.7$ & $26.7 \pm 2.4$ \\
\segrpo{} (self-context) & $80.4 \pm 2.3$ & $23.4 \pm 2.1$ \\
\midrule
OPD (self-context) & $78.3 \pm 2.4$ & $21.6 \pm 2.0$ \\
\bottomrule
\end{tabular}
\vspace{4pt}
\caption{Agentic task final success rate (\%) with Qwen2.5-7B-Instruct on ALFWorld and ScienceWorld. Each entry is the mean $\pm$ sample standard deviation over ten independent end-to-end training-and-evaluation runs. All methods are evaluated on the full validation set of each benchmark under the same decoding configuration. Per-run final-checkpoint success is computed as avg@1 with one full multi-turn rollout per task and no tool, retry, or admissible-action assistance. The \segrpo{} (DeepSeek-V4) row uses external DeepSeek-V4 verified rollouts as proposals; the \segrpo{} (self-context) row uses the self-context proposal source with the failure-lesson memory of Appendix~\ref{app:self-context}. The OPD row uses the same self-context proposals but replaces forward-KL consolidation with on-policy distillation, confirming that forward-KL is the stronger consolidation objective in the agentic setting as well.}
\label{tab:alfworld-final-results}
\end{table}

\begin{table}[!ht]
\centering
\begin{tabular}{lcccc}
\toprule
Task type (hardest first) & Base (step $0$) & GRPO & TREK & $\Delta$ \\
\midrule
Examine in Light & $50.0$ & $56.2 \pm 6.2$ & $68.8 \pm 3.2$ & $+12.6$ \\
Heat \& Place & $21.1$ & $59.5 \pm 2.4$ & $78.6 \pm 3.1$ & $\mathbf{+19.1}$ \\
Pick Two \& Place & $0.0$ & $63.2 \pm 0.6$ & $71.1 \pm 2.9$ & $+7.9$ \\
Cool \& Place & $8.7$ & $65.0 \pm 5.0$ & $72.5 \pm 2.5$ & $+7.5$ \\
Pick \& Place & $37.9$ & $91.2 \pm 0.4$ & $88.2 \pm 2.9$ & $-3.0$ \\
Clean \& Place & $12.9$ & $92.3 \pm 3.8$ & $100.0 \pm 0.0$ & $+7.7$ \\
\midrule
Overall & $18.8$ & $75.8 \pm 2.1$ & $82.8 \pm 2.7$ & $+7.0$ \\
\bottomrule
\end{tabular}
\vspace{4pt}
\caption{Per-task-type ALFWorld success rate (\%) for the DeepSeek-V4 \segrpo{} variant across all six task types, ordered by GRPO baseline success (lowest first). The Base column is the untrained deployment student at step $0$; the GRPO and TREK columns are reported as mean $\pm$ sample standard deviation over the same ten independent end-to-end runs as the DeepSeek-V4 row in Table~\ref{tab:alfworld-final-results}. The lowest-baseline task types, Examine in Light and Heat \& Place, exhibit the largest gains from TREK, whereas the near-saturated Pick \& Place and Clean \& Place have little remaining headroom.}
\label{tab:alfworld-bytask}
\end{table}

The agentic results show the same pattern in long-horizon action settings. On ALFWorld, direct GRPO reaches $75.8$; external DeepSeek-V4 proposals raise success to $82.8$, while the self-context proposal source reaches $80.4$ without changing the deployable student interface. On ScienceWorld, the gains are even more pronounced in relative terms: GRPO reaches $12.5$, while \segrpo{} with DeepSeek-V4 proposals more than doubles success to $26.7$ and the self-context variant reaches $23.4$. This is consistent with the math results: using verified proposals as a support-expansion step on hard prompts before GRPO refinement resumes yields clear gains over ordinary on-policy GRPO, and these gains carry over to long-horizon action settings even when the proposal source is the same model with additional inference-time context rather than an external teacher.

Most importantly, the per-task breakdown in Table~\ref{tab:alfworld-bytask} shows that these gains concentrate on the harder task types rather than spreading uniformly across the benchmark. Ranking task difficulty by the GRPO baseline success rate, the lowest-baseline types are Examine in Light and Heat \& Place at $56.2\%$ and $59.5\%$, well below the near-saturated Pick \& Place and Clean \& Place at $91.2\%$ and $92.3\%$. These low-baseline types show the largest improvements from \segrpo{}: Heat \& Place gains $+19.1$ points and Examine in Light gains $+12.6$ points over the GRPO baseline, while the remaining lower-baseline types, Pick Two \& Place and Cool \& Place, gain $+7.9$ and $+7.5$. By contrast, the already-strong Pick \& Place has essentially no headroom and moves within noise ($-3.0$), while Clean \& Place is pushed to the ceiling ($+7.7$, reaching $100.0\%$). This directly supports the central claim that support-expansion proposals help most where unaided exploration is weakest: the task types on which the deployment student is least able to discover successful trajectories are precisely the ones that benefit most.

\begin{figure}[!ht]
    \centering
    \begin{minipage}[t]{0.48\linewidth}
        \centering
        \includegraphics[width=\linewidth]{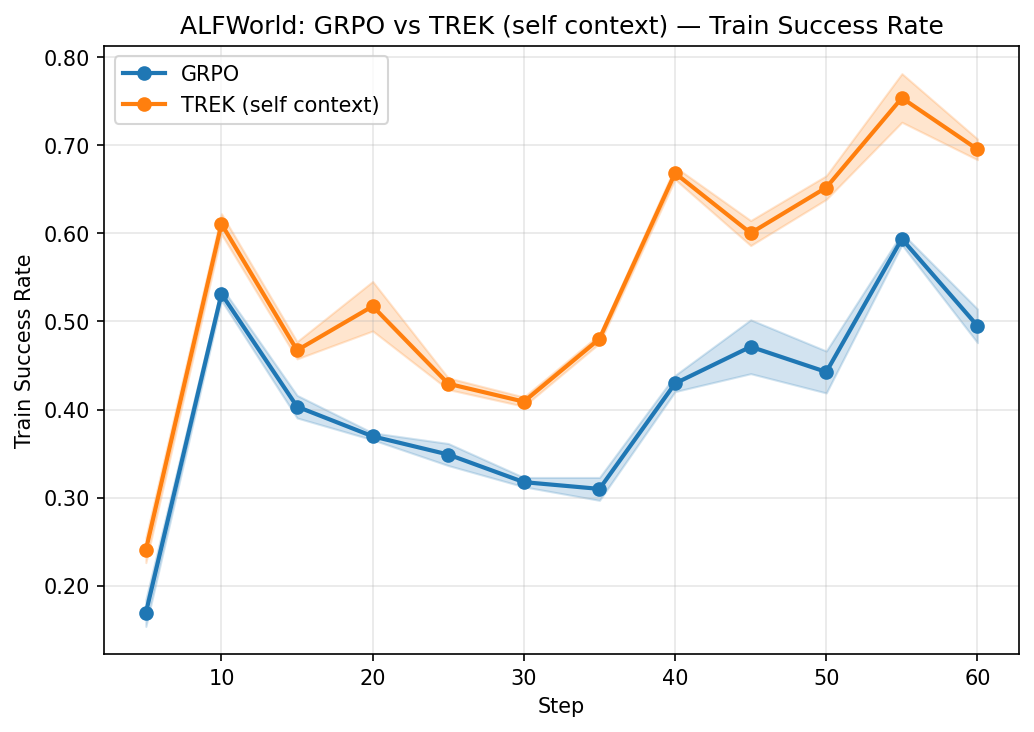}
    \end{minipage}\hfill
    \begin{minipage}[t]{0.48\linewidth}
        \centering
        \includegraphics[width=\linewidth]{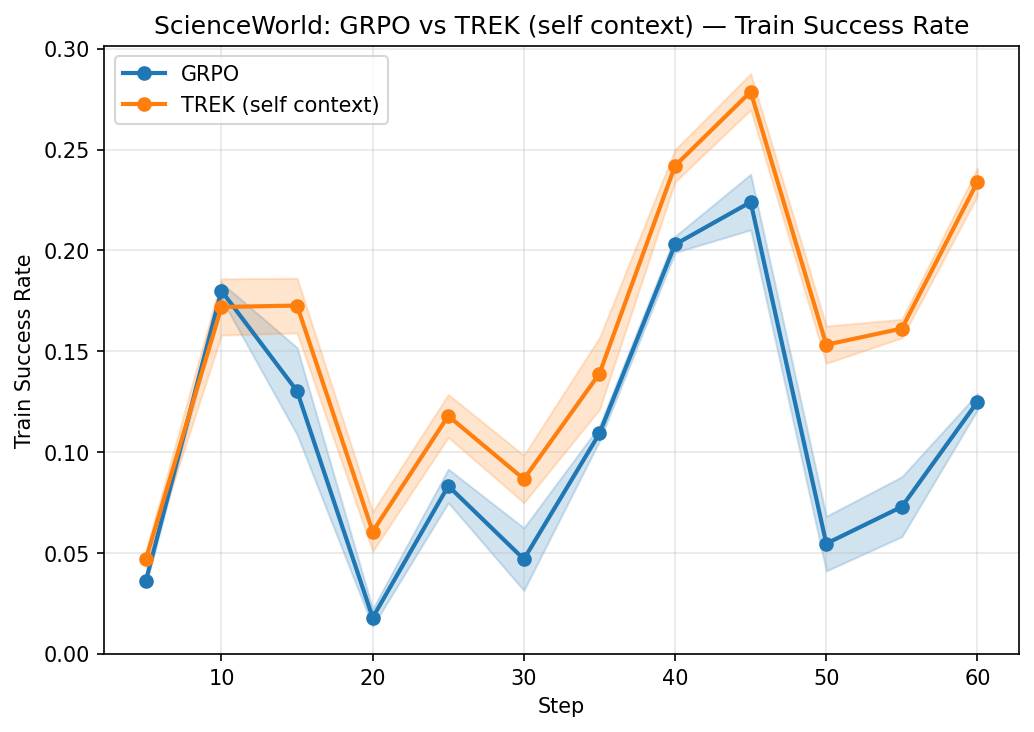}
    \end{minipage}
    \caption{Training success rate over optimization steps on ALFWorld (left) and ScienceWorld (right). In both environments, \segrpo{} (self-context) reaches high success rates much earlier than direct GRPO, demonstrating that proposal-based support expansion accelerates learning on hard tasks rather than merely raising the asymptotic ceiling.}
    \label{fig:train-curves}
\end{figure}

We note that direct GRPO can eventually reach comparable overall success rates (approximately $85\%$) if trained for substantially longer. The primary advantage of \segrpo{} is therefore not the asymptotic ceiling but \emph{training efficiency on hard tasks}: by consolidating verified proposals early, \segrpo{} pulls success rate on the hardest task types up much earlier in training, reaching strong performance in a fraction of the steps that unaided GRPO requires to discover the same solution modes through random exploration alone. Figure~\ref{fig:train-curves} illustrates this acceleration effect clearly across both agentic benchmarks. On ALFWorld, \segrpo{} (self-context) already exceeds $60\%$ train success rate by step $20$, while GRPO remains below $50\%$ at the same point and requires roughly $5\times$ more steps to reach the same level. On ScienceWorld, the same pattern holds: \segrpo{} opens a consistent gap over GRPO in the early training phase, with the separation emerging by step $10$ and persisting throughout training. The gap is most pronounced in the early phase of training: the forward-KL consolidation step injects verified solution modes into the student's support before RL begins, so the student enters the GRPO phase with hard-task trajectories already sampleable rather than needing to discover them from scratch through random exploration. This early acceleration is practically important because agentic rollouts are expensive, and reducing the number of training rounds needed to cover hard-task modes directly reduces the total environment-interaction budget.

\subsection{Efficiency and Implementation Details}

All pass-rate estimates use held-out rollouts from the current checkpoint rather than reuse the same samples that drive the training update. For math, answer extraction and verification are identical across methods. For ALFWorld, the student is trained on action trajectories and evaluated in the environment without extra planning context, admissible-action pruning, backtracking, or verifier-guided retry. The full ALFWorld and ScienceWorld train-and-evaluate pipelines are repeated over ten independent runs so that reported success rates reflect end-to-end training variance rather than a single evaluation seed.

\section{Related Work}

We situate \segrpo{} relative to teacher-informed RL and self-improvement methods along one main axis: what the teacher signal is used for. Prior distillation and OPD-style recipes usually improve supervision, credit assignment, or reliability over behaviors already sampled by the student. In contrast, \segrpo{} uses teacher or self-context rollouts as verified proposals for missing-support prompts, filters them by prompt difficulty and student-proximal trajectory reachability, briefly consolidates them with forward KL, and then returns to ordinary on-policy GRPO. Because the method consumes only verified output trajectories, the proposal source can be an external black-box teacher, a white-box teacher, or the same model with additional inference-time context.

\paragraph{Reasoning RL with verifiable rewards.}
GRPO and related outcome-supervised policy optimization methods improve reasoning models by optimizing verifiable reward over sampled responses \citep{shao2024deepseekmath}. Recent analyses suggest that current RLVR often improves small-$k$ performance without eliciting fundamentally new reasoning patterns beyond the base model's large-$k$ coverage \citep{yue2025does}. Related reasoning fine-tuning recipes also combine reinforcement signals or reference solutions to improve hard-problem learning \citep{luong2024reft,wu2026regft}. Our work preserves GRPO as the final optimizer but targets a different failure mode: missing support over reward-bearing trajectories.

\paragraph{On-policy distillation and self-distillation.}
On-policy distillation reduces exposure mismatch by supervising student rollouts rather than fixed teacher trajectories \citep{agarwal2024onpolicy,gu2024minillm}. Related extensions study token weighting and token-level teachability, privileged-information distillation, competence-adaptive supervision, input-specific credit assignment, iterative self-policy distillation, binary-reward-to-dense self-distillation, reliability-aware OPD variants, and sparse-to-dense reward shaping \citep{xu2026paced,xu2026tip,wang2026teachability,shen2026credit,sang2026beyond,penaloza2026pidistill,sang2026crisp,he2026sdzero}. SRPO \citep{li2026srpo} unifies GRPO and self-distillation by routing correct samples to GRPO and failed samples to logit-level self-distillation within the same on-policy group. These methods primarily use teacher information to improve supervision or credit assignment on trajectories that are already available. \segrpo{} differs in three ways: (i) it routes at the \emph{prompt level} to target prompts where the rollout pass rate is very low rather than routing individual samples within an already-successful group; (ii) it only requires verified output trajectories, not teacher logits, making it compatible with black-box proposal sources; and (iii) its forward-KL consolidation objective directly optimizes mode coverage over verified proposal modes, which our OPD ablation (Tables~\ref{tab:qwen3-math-results} and~\ref{tab:alfworld-final-results}) shows is a stronger support-expansion signal than OPD-style supervision on the same trajectories.

The recent OPD design space is broader than the subset we rely on directly here, including context distillation, black-box OPD, and survey-style syntheses \citep{song2026opdsurvey,hou2026uniopd,li2026rethinkopd,ye2026opcd,ye2026blackboxopd}. This line is complementary: token teachability and OPD-style methods ask how to allocate supervision more effectively within student-reachable behavior, whereas \segrpo{} targets the earlier failure mode in which the student does not reliably reach reward-bearing modes in the first place.

\paragraph{Same-model self-context and self-improvement.}
Several lines of work improve reasoning by using the same base model with additional inference-time context or computation, including self-consistency, self-bootstrapping, iterative self-feedback, verbal reflection, process verification, self-improvement, and search over intermediate thoughts \citep{wang2022selfconsistency,zelikman2022star,shinn2023reflexion,madaan2023selfrefine,yao2023tree,lightman2023let,huang2022selfimprove}. Closely related work also studies different aspects of whether teacher explanations, reasoning traces, progressively supported references, context-efficient adaptation and context management, or robustness under distribution drift can make smaller or deployed models more capable under practical resource constraints \citep{li2022explanations,hsieh2023distilling,liu2025trainingwheels,hu2026codes,shen2026efficiencyfrontier,xin2026driftguardsafetyawaremultimonitordetection}. Our setting is different in two ways. First, we use these mechanisms as proposal generators for training rather than as permanent test-time procedures. Second, we retain only verified top-$r$ student-proximal trajectories, then fold them back into the unaided policy through a short forward-KL phase.

\paragraph{Classical distillation and sequence-level transfer.}
Classical knowledge distillation and sequence-level distillation already show that a student can learn effectively from teacher outputs without direct access to teacher internals \citep{hinton2015distilling,kim2016sequence}. Our method inherits that output-level compatibility, but differs in objective and setting. We do not use teacher sequences as a general-purpose imitation target; instead, verified trajectories are used selectively to expand support on routed hard prompts before the student returns to ordinary on-policy RL. This makes the method closer to targeted support expansion than to standard sequence-level compression.

\paragraph{Curriculum, self-paced learning, and routing.}
Curriculum learning and self-paced learning choose which examples to emphasize based on difficulty or current competence \citep{bengio2009curriculum,kumar2010selfpaced}. Our prompt routing has a related flavor, since it uses current student pass rate together with verified-proposal availability to decide which low-pass-rate prompts receive proposal learning. The key difference is that routing here is not only about ordering or weighting examples: it also determines when to invoke an external or context-augmented proposal mechanism and when to run a short forward-KL support-expansion phase.

\paragraph{Teacher-guided RL updates.}
Methods such as RLSD and reinforcement-aware distillation integrate teacher information into token-level credit assignment or trust-region updates during RL \citep{yang2026rlsd,zhang2026reinforcementaware}. Their focus is how teacher signals should modulate reward-driven updates on student rollouts. Our focus is earlier in the pipeline: when no reward-bearing trajectories are sampled, the first problem is not finer credit assignment but exploration recovery.

\paragraph{Exploration and tool-augmented reasoning.}
Many reasoning systems improve performance by adding computation at inference time, including self-consistency, search, execution feedback, reflection, and acting in environments \citep{wang2022selfconsistency,yao2023react,shinn2023reflexion,yao2023tree}. \segrpo{} uses the same family of mechanisms in training, but only as training-time proposal generators whose outputs are filtered for reachability and then transferred back into the unaided student.

\section{Discussion and Limitations}

Several practical considerations are worth noting. Like all verifier-based training methods, the quality of proposal filtering depends on the verifier: stronger checks give more informative selection. The reachability proxy (trimmed NLL) is effective for ranking but does not capture fine-grained learnability, and it remains sensitive to verbosity and surface form. The method also adds teacher queries and verifier calls on top of ordinary GRPO; more adaptive scheduling could reduce this overhead at the cost of additional engineering.

The verified-only rule is a conservative choice rather than a fundamental requirement. Teacher trajectories that fail the verifier can still expose useful reasoning patterns, alternative decompositions, or subgoal structures that broaden the student's exploration. The main reason we restrict to verified trajectories is practical: for the student scales studied here, unverified teacher trajectories are often far from the student's current support, making consolidation unstable. With larger student capacity, relaxing the verifier gate to include near-miss teacher trajectories is a natural extension.

\section{Conclusion}

We proposed a broader view of distillation as exploration support expansion and instantiated it in \segrpo{}, a simple routed training procedure for hard prompts. The core organizing idea is to separate prompt-level proposal availability from trajectory-level student-proximity selection: first decide where extra proposal generation is justified, then choose the verifier-passing successes that the current student is most likely to internalize. The resulting recipe uses verified proposals only where ordinary student sampling has very low pass rate, retains the top-$r$ verified proposals by current student likelihood, applies a short forward-KL warm-start to move those solution modes into the student's support, and then returns to ordinary on-policy GRPO refinement. Under this view, distillation and RL play different roles: distillation expands where the student can search, while GRPO determines what the student ultimately learns to do under verifiable reward. The same logic is not specific to one benchmark family: it applies across answer-verifiable reasoning tasks such as mathematics and across agentic tasks where exploration failures appear as missing action trajectories or subgoal sequences. Although we instantiate the idea with GRPO, the same routed proposal-learning recipe should extend naturally to related post-training frameworks such as DAPO and GSPO \citep{yu2025dapo,gspo2025}, and other group-based or on-policy objectives with the same exploration bottleneck. The same staged decomposition may also serve as a natural ingredient for continual skill acquisition: when the model encounters new tasks, context-conditioned verified proposals can introduce successful trajectories for the new regime, after which the forward-KL stage consolidates them into the student's own support before on-policy refinement stabilizes them under the deployable interface; we leave empirical validation of this extension to future work. We hope this framing encourages a more general perspective on how teacher information can help on-policy reasoning systems beyond imitation and dense credit assignment alone.

\section*{Acknowledgments}

We thank Li Dong and Tianzhu Ye for their valuable discussions.

\bibliographystyle{unsrtnat}
\bibliography{references}

\appendix

\section{MATH-500 Results}
\label{app:math500}

For completeness, Table~\ref{tab:qwen3-math500-results} reports the MATH-500 numbers omitted from the main text. Direct GRPO is already near-saturated on MATH-500 (roughly $88$--$90\%$ at the 8B and 14B scales), so the headroom for support-expansion proposals is small; \segrpo{} still improves over direct GRPO at every scale, but the gains are modest compared with the harder AIME benchmarks in Table~\ref{tab:qwen3-math-results}. The MATH-500 ablation breakdown is omitted because the saturated regime provides little discriminative signal among the control conditions.

\begin{table}[h]
\centering
\begin{tabular}{llc}
\toprule
Stage & Model & MATH-500 \\
\midrule
Direct GRPO & Qwen3-1.7B & $76.2 \pm 1.1$ \\
Direct GRPO & Qwen3-8B & $88.3 \pm 0.9$ \\
Direct GRPO & Qwen3-14B & $89.5 \pm 0.8$ \\
\midrule
\segrpo{} (DeepSeek-V4) & Qwen3-1.7B & $79.6 \pm 1.2$ \\
\segrpo{} (DeepSeek-V4) & Qwen3-8B & $91.5 \pm 1.1$ \\
\segrpo{} (DeepSeek-V4) & Qwen3-14B & $92.7 \pm 1.9$ \\
\midrule
\segrpo{} (self-context) & Qwen3-1.7B & $78.5 \pm 1.3$ \\
\segrpo{} (self-context) & Qwen3-8B & $90.1 \pm 1.2$ \\
\segrpo{} (self-context) & Qwen3-14B & $91.8 \pm 1.2$ \\
\bottomrule
\end{tabular}
\vspace{4pt}
\caption{MATH-500 results across Qwen3 scales (avg@16, \%). The reported \segrpo{} results use verified DeepSeek-V4 rollouts as proposal trajectories, then consolidate the selected trajectories into the Qwen3 deployment model through a short forward-KL proposal-learning phase before GRPO refinement resumes. The self-context rows at the bottom are the matched same-model context variant, where the proposal source is the deployment model itself with additional inference-time context. Direct GRPO is already near-saturated on MATH-500, so the gains are smaller than on the AIME benchmarks reported in the main text.}
\label{tab:qwen3-math500-results}
\end{table}

\section{Experimental Setup Details}
\label{app:setup-details}

We list here the concrete settings of the routing and selection protocol described in the main text. Every training round estimates $\ps(x)$ from $K=16$ unaided student rollouts, with hardness gate $\tau_{\mathrm{low}}=1/8$. For each hard candidate the proposal source is queried $M=4$ times, and the top $r=2$ verifier-passing proposals by trimmed length-normalized NLL form the consolidation set. The trim fractions are $(\alpha,\beta)=(0.10,0.02)$ as quantile trims (lowest $10\%$ and highest $2\%$ of token losses removed before averaging). The proposal-learning window is one epoch over the accumulated retained proposals for that round before the policy continues in the ordinary GRPO stream. The 2$\times$ rollout-budget GRPO baseline uses 32 unaided rollouts per prompt and no teacher proposals. The ALFWorld experiments use Qwen2.5-7B-Instruct as the deployable student backbone.

Table~\ref{tab:math-config} gives the math configuration used for the MATH-500 and AIME evaluations. We use the DAPO-Math-17K training pool for verifier-based math post-training and evaluate only on held-out benchmarks. The GRPO hyperparameters follow the same math post-training protocol across direct-GRPO, proposal-learning, and refinement comparisons; the TREK-specific routing parameters are the $K$, $M$, and $r$ values described above.

\begin{table}[h]
\centering
\small
\begin{tabular}{p{0.42\linewidth}p{0.50\linewidth}}
\toprule
Setting & Value \\
\midrule
Training pool & DAPO-Math-17K \\
Evaluation sets & MATH-500, AIME 2024, AIME 2025 \\
Framework & VERL / HybridFlow \\
Optimizer & AdamW \\
Learning rate & $1\times10^{-7}$ \\
GRPO epochs & 10 \\
GRPO mini-batch / micro-batch & 8 / 1 per GPU \\
GRPO max prompt / response tokens & 3072 / 16384 \\
Proposal max prompt / response tokens & 2048 / 16384 \\
Proposal epochs & 1 \\
Proposal mini-batch / micro-batch & 64 / 4 per GPU \\
Sampling temperature / top-$p$ / top-$k$ & 1.0 / 1.0 / -1 \\
Clip ratio / gradient clip & 0.2 / 1.0 \\
KL coefficient & $5\times10^{-4}$ \\
Evaluation protocol & avg@16 \\
2$\times$ rollout baseline & 32 unaided samples per prompt; no teacher proposals \\
Precision & bfloat16 \\
Rollout engine / TP size & sglang / 16 for GRPO; sglang / 2 for proposal learning \\
Compute & $2\times 8\times$ NVIDIA H200 \\
\bottomrule
\end{tabular}
\vspace{4pt}
\caption{Math implementation configuration.}
\label{tab:math-config}
\end{table}

Table~\ref{tab:alfworld-config} summarizes the ALFWorld-specific implementation configuration. We use the text-only ALFWorld environment rather than embodied visual observations, and evaluate the final policy without extra planning context or admissible-action filtering.

\begin{table}[h]
\centering
\small
\setlength{\tabcolsep}{4pt}
\begin{tabular}{p{0.42\linewidth}p{0.50\linewidth}}
\toprule
Setting & Value \\
\midrule
Evaluation set & ALFWorld evaluation split; 128 instances \\
Framework & VERL / HybridFlow \\
Optimizer & AdamW \\
Learning rate & $7\times10^{-7}$ \\
GRPO epochs (total) & 150 \\
PPO mini-batch / micro-batch & 256 / 32 per GPU \\
Train rollout batch & 16 prompts $\times$ 8 group\\
Max prompt / response tokens & 2048 / 512 per turn \\
Environment max steps & 50 action turns per episode \\
Sampling (eval) & temperature 0.4 \\ 
Clip ratio / gradient clip & 0.2 / 1.0 \\
KL coefficient (loss) & 0.001 \\
Precision & bfloat16 \\
Rollout engine / TP size & vLLM / 2 (GPU memory utilization 0.5) \\
\bottomrule
\end{tabular}
\vspace{4pt}
\caption{ALFWorld implementation configuration.}
\label{tab:alfworld-config}
\end{table}

We implement the training pipeline on top of the HybridFlow/VERL post-training stack \citep{sheng2024hybridflow}, so that GRPO rollouts, proposal generation, verifier evaluation, and the short forward-KL update share model checkpoints, decoding settings, and verifier outputs inside a single distributed loop. Each round, the student first generates unaided rollouts for routing; selected prompts are sent to a proposal worker (a larger teacher model or the same base model with additional inference-time context), and verified proposal trajectories are logged with prompt identifiers and student likelihood statistics so that top-$r$ selection and the forward-KL update use exactly the same prompt snapshot. The forward-KL phase is a teacher-forced negative log-likelihood pass on the retained proposal set with the same tokenizer and response formatting as the student, after which the loop returns to ordinary GRPO without forcing an extra resampling pass.

\subsection{Optional Pipelined Scheduling for Faster Proposal Learning}
\label{app:pipelined-schedule}

The scheduler need not block ordinary GRPO while waiting for proposal rollouts. As a possible acceleration strategy, TREK can be run as a delayed two-queue pipeline. At training round $t$, the student first generates unaided rollouts for the current prompt batch. Prompts above the hard-prompt cutoff are used immediately for ordinary GRPO updates. Prompts with low pass rate are recorded as hard candidates and placed into a teacher-proposal queue, rather than forcing the current student update to wait for the proposal source.

At a later round $t+k$, once teacher proposals for earlier hard prompts are available, the verifier filters them and the current student re-scores the verifier-passing trajectories using $\nll(y_T\mid x)$. Re-scoring with the current student is important because the policy may have changed since the prompt was first mined as hard; stale proposal rankings can otherwise select trajectories that are no longer student-proximal. We then keep the top-$r$ verifier-passing proposals, apply the short forward-KL update to those cached proposal examples, and return the corresponding prompts to the ordinary GRPO sampling pool. No immediate resampling pass is required; subsequent on-policy rollouts are collected when those prompts next appear in the normal training stream.

This pipelined form preserves the staged interpretation of the loss, but it is not required by the algorithm. Prompts above the hard-prompt cutoff receive $\mathcal{L}_{\mathrm{GRPO}}$ immediately from student rollouts. Cached hard prompts receive $\lambda\mathcal{L}_{\mathrm{FKL}}$ only after verified top-$r$ proposals are ready, and their subsequent GRPO updates are computed from later on-policy student rollouts rather than from the teacher trajectories themselves. Thus the expression $\lambda\mathcal{L}_{\mathrm{FKL}}+\mathcal{L}_{\mathrm{GRPO}}$ should be read as the prompt's training path across the proposal and refinement phases, not as requiring a single synchronous minibatch or an immediate post-FKL resampling step.

The same pipeline can reduce overhead in three simple ways. First, proposal generation is invoked only for low-$\ps$ prompts and can run asynchronously with student GRPO on prompts above the hard-prompt cutoff. Second, teacher generation can early-stop once $r$ verifier-passing proposals have been found, up to the maximum budget $M$. Third, if a cached prompt is no longer hard when its proposals become available, the forward-KL update can be skipped and the prompt can be handled by ordinary GRPO.

\section{Self-Context Variant: Failure-Lesson Memory}
\label{app:self-context}

The self-context variant uses the same Qwen3 deployment checkpoint as the proposal generator, but augments its prompt with a short natural-language memory of \emph{failure lessons} distilled from the model's own previously verifier-rejected math attempts. The lesson set is built once before training: we collect a large pool of unaided rollouts on math prompts, take only the verifier-rejected trajectories, and ask the same Qwen3 model to summarize recurring error patterns into compact, transferable rules. We then deduplicate and lightly curate the resulting set. In practice the memory contains roughly forty rules covering algebraic manipulation errors, casework omissions, modular arithmetic misuse, geometric misreadings, optimization boundary cases, counting conventions, and final-answer mismatches.

At proposal generation time, this memory is prepended to the math problem as additional context. The same Qwen3 checkpoint then samples solutions under this enriched prompt, and only verifier-passing trajectories are kept. The retained trajectories are scored under the unaided student likelihood and consolidated through the same top-$r$ selection and short forward-KL phase as the DeepSeek-V4 variant; the failure-lesson memory is used only at proposal generation time, never at deployment evaluation.

For illustration, we reproduce a representative excerpt from the failure-lesson memory below, formatted as it is actually prepended to the math problem at proposal generation time. The full memory contains roughly forty such rules; only seven are shown here to convey the granularity and style.

\begin{quote}
\small\ttfamily
Before solving, recall these lessons from previous failed attempts:\\[2pt]
-- Re-read what the problem actually asks. Make sure the final boxed value matches the requested quantity (sum, count, remainder, probability, etc.), not an intermediate variable.\\
-- List hidden constraints before computing: integer, positive, nonzero, distinct, domain of a substitution. Filter algebraic candidates against these constraints before answering.\\
-- After deriving a candidate answer, verify it in the original problem (substitute back, check small cases, or check geometric conditions). Do not stop at the first plausible value.\\
-- If you square both sides of an equation, every candidate must be substituted back into the original equation; squaring creates extraneous roots.\\
-- Never divide both sides by an expression like x, x-y, or sin(theta) without separately handling the case where that expression is zero.\\
-- In counting and probability, decide ordered vs.\ unordered, labeled vs.\ unlabeled, independent vs.\ dependent before applying a formula. Use complement counting when ``at least one'' / ``not all'' appears.\\
-- For modular arithmetic on a\textasciicircum n mod m, do not blindly apply Fermat/Euler. Verify gcd(a, m) = 1 first; otherwise compute the cycle of a modulo m directly.\\[2pt]
Now solve the following problem step by step.
\end{quote}

\end{document}